# High-level programming and control for industrial robotics: using a hand-held accelerometer-based input device for gesture and posture recognition

Pedro Neto[†], J. Norberto Pires[†], A. Paulo Moreira[††]

[†]Mechanical Engineering Department, University of Coimbra, Coimbra, Portugal
[††]Department of Electrical and Computer Engineering, University of Porto, Porto, Portugal

*Abstract*: Today, most industrial robots are still programmed using the typical teaching process, through the use of the robot teach pendant. This paper presents a robotic system that allows users, especially non-expert programmers, to instruct and program a robot just showing it what it should do, and with a high-level of abstraction from the robot language. This is done using the two most natural human interfaces (gestures and speech), a force control system and several code generation techniques.

Special attention will be given to the recognition of gestures, where the data extracted from a motion sensor (3-axis accelerometer) embedded in the Wii Remote Controller was used to capture human hand behaviours. Gestures (dynamic hand positions) as well as manual postures (static hand positions) are recognized using a statistical approach and Artificial Neural Networks (ANNs).

Several experiments are done to evaluate the proposed system in a non-controlled environment and to compare its performance with a similar approach, which instead of gestures uses a manual guidance system based on a force control strategy. Finally, different demonstrations with two different robots are presented, showing that the developed system can be customized for different users and robotic platforms.

1. Introduction

1.1 Motivation

Programming an industrial robot by the typical teaching method, through the use of the robot teach pendant is a tedious and time-consuming task that requires some technical expertise. In industry, this type of robot programming can be justified economically only for production of large lot sizes. Hence, new approaches to robot programming are required.

Contrary to the highly intelligent robots described in science fiction, most current industrial robots are "non-intelligent" machines that work in a controlled and well known environment. Generally, robots are designed, equipped and programmed to perform specific tasks, and thus, an unskilled worker will not be able to re-program the robot to perform a different task.

The goal is to create a methodology that helps users to control and program a robot with a high-level of abstraction from the robot language. Making a demonstration in terms of high-level behaviours (using gestures, speech, etc.), the user should be able to demonstrate to the robot what it should do, in an intuitive way. This type of learning is often known as programming by demonstration (PbD). Several approaches for PbD have been investigated, using different input devices, manipulators and learning strategies [1-3]. The demand for new and natural human-machine interfaces (HMIs) has been increasing in recent years, and the field of robotics has followed this trend [4]. Speech recognition is seen as one of the most promising interfaces between humans and machines, since it is probably the most natural and intuitive way of communication between humans. For this reason, and given the high demand for intuitive HMIs, automatic speech recognition (ASR) systems have had a great interest shown in them in the last few years. Today, these systems present good performance and robustness, allowing, for example, the control of industrial robots in an industrial environment (in the presence of surrounding noise) [5]. Gestures are another natural form of communication between humans. In the robotics field, work has been done in order to identify and recognize human gestures. There are various ways to capture human gestures, using vision-based interfaces [6-7], motion capture sensors [2], using the combination of both (a vision system and a data glove) [1], or using finger gesture recognition systems based on active tracking mechanisms [8].

Accelerometer-based gesture recognition has become an emerging technology, providing new possibilities to interact with machines like robots. Some accelerometer-based input devices have been developed to work as a flexible interface for modern consumer electronic products. In order to recognize gestures, the acceleration data extracted from these devices has been used as input for Artificial Neural Networks (ANN) models [9] and Hidden Markov Models (HMM) [10]. In other work, accelerometers and surface EMG sensors are used synchronously to detect hand movements [11]. An interesting approach presents a method of recognition of lower limb movements using a 3-axis accelerometer and ANNs [12].

Although, several systems use HMM for accelerometer-based gesture recognition [13] and ANNs have been applied in a wide range of situations, such as in the recognition of gestures for sign language [14] and vision-based gesture recognition systems [15]. In many research works, ANNs have produced very satisfying results and have proven to be efficient for classification problems.

Notwithstanding the above, due to the specific characteristics of an industrial environment (colours, non-controlled sources of light and infrared radiation, etc.) it remains difficult to apply such systems, especially when certain types of infrared and vision based systems are used. The reliability of technologies is also an important issue as many systems have not yet reached industrial usage. Given the above, the teach pendant continues to be the common robot input device that gives access to all functionalities provided by the robot and the controller (jogging the manipulator, producing and editing programs, etc.). In the last few years, the robot manufacturers have made great efforts to make user-friendly teach pendants, implementing ergonomic design concepts, more intuitive user interfaces such as icon-based programming [16], colour touch screens with graphical interfaces, a 3D joystick (ABB) [17], a 6D mouse (KUKA) [18-19], and developing a wireless teach pendant (COMAU) [20]. Nevertheless, it is still difficult for an untrained worker to operate with a robot teach pendant. The teach pendants are not intuitive to use and require a lot of user experience, besides being big, and heavy [21]. It is interesting to note that in the opinion of many robot programmers, the cable that connects the teach pendant to the robot controller is one of the biggest drawbacks of the equipment.

Several studies have been done to investigate intuitive ways to move and teach robots, using input devices, such as joysticks [22] or digital pens [23]. Due to its low price and specific characteristics (see section 2.2), the Wii Remote controller was selected to be the input device of our system, a wireless device with motion and infrared sensing capabilities.

1.2 Objectives

The purpose of this work was to develop a system to teach and program industrial robots by "natural means", using gestures and speech. Gestures can be considered a natural communication channel, which has not yet been fully utilized in human-robot interaction. Therefore, the aim is to increase the use of these systems in the robotics field. The game controller Wii Remote is used to capture human hand behaviours, manual postures (static hand positions) and gestures (dynamic hand positions). The information collected from the Wii Remote (motion data) will be used to jog the robot. These motion data extracted from the 3-axis accelerometer embedded in the Wii Remote, are used as input to a statistical model and an ANN algorithm previously trained. The outputs of this statistical and ANN algorithm is the recognized gestures and postures that are then used to control a robot in the way required. The developed system incorporates ASR software that allows the user to manage the cell, acting on the robot and on the code generation system. Also included is a force control system to avoid excessive contact forces between the robot tool and workpiece. This system also detects obstacles and avoids collisions during the robot operation.

Several experiments were performed to verify the viability of this system in a non-controlled environment (industrial environment) and to compare its performance with others, especially one similar system that instead of gestures uses a manual guidance system based on a force control strategy. Different practical tests (pick-and-place, write on a sheet of paper, and move the robot to different poses in the presence of obstacles) with two different robots (MOTOMAN and ABB) are presented, showing that the developed system can be customized for different users and robots. Possible future research directions are discussed and conclusions made.

## 2. Experimental setup

2.1 System description

The demonstration cell (Fig. 1) is composed of an industrial robot (HP6 equipped with the NX100 controller, MOTOMAN, Japan), a Wii Remote controller (Nintendo, Japan) to capture human hand behaviours, a standard headset microphone to capture the user voice, a force/torque (F/T) sensor (85M35A-I40, JR3, USA), and a computer running the application that manages the cell (Fig. 2).

Using gestures, the user moves the robot to the desired pose, and then, through voice commands, the user can save the corresponding robot configuration and generate the robot code. The user can then move the robot to another point, and so on, until it reaches the desired robot path.

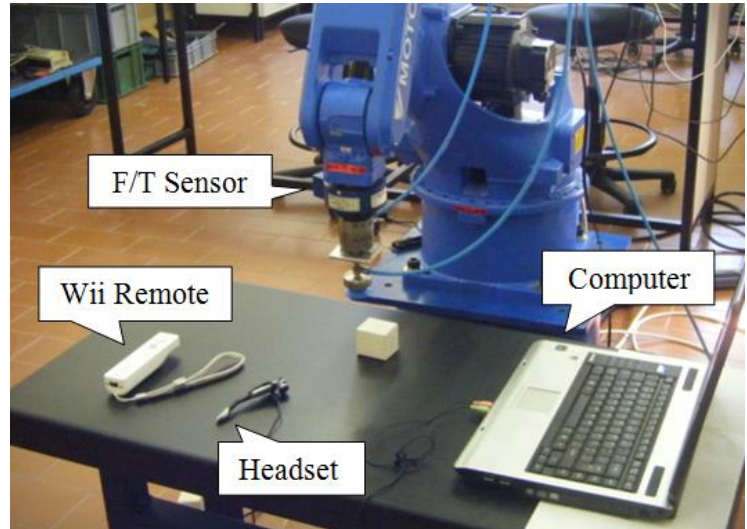

Fig. 1. The robotic cell is basically composed of an industrial robot, a F/T sensor, two input devices (Wii Remote and headset), and a computer running the application that manages the cell.

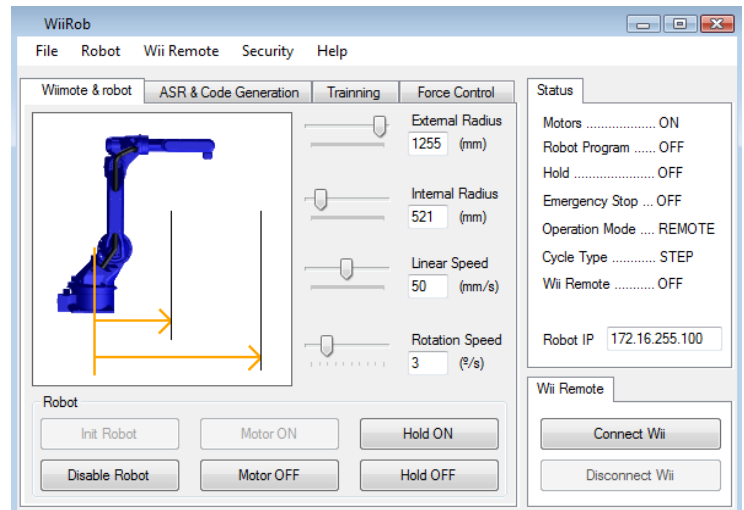

Fig. 2. Application interface developed using Microsoft Visual C#.

The application receives data from the Wii Remote, interprets the received data and sends commands to the robot. For this purpose, the MotomanLib, a Data Link Library was created in our laboratory to control and manage the robot remotely via Ethernet (Fig. 3). The application has incorporated ASR software that recognizes the voice commands received from the headset and, depending on the commands received acts on the robot or on the code generation system that is embedded in the application.

An active X component named JR3PCI was used to communicate with the F/T sensor [24], allowing the application to continuously receive feedback from the F/T sensor. If any component of force or torque exceeds a set value, a command is sent, making the Wii Remote vibrate (tactile feedback), alerting the user. This is a way of providing feedback about the cell state to the user, in addition to the sound feedback (alert sounds and text to speech (TTS) software that reports the cell state and occurrences). Finally, the application also includes a section to train the statistical model and the ANN.

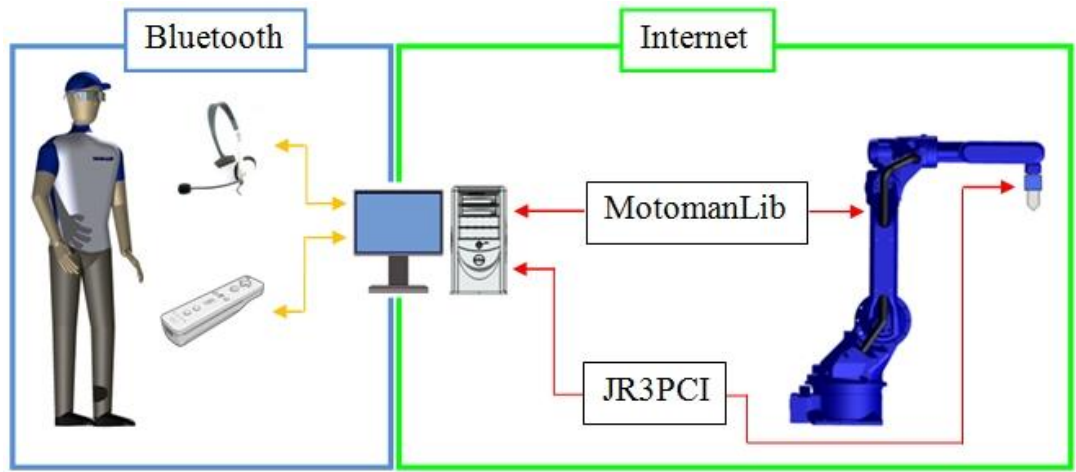

Fig. 3. Communication and system architecture. The input devices work without wires (via Bluetooth), giving a greater freedom to the user.

2.2 The Wii Remote

The demand for new interaction systems to improve the game experience has led to the development of new devices that allow the user to feel more immersed in the game. In contrast to the traditional game pads or joysticks, the Wii Remote allows users to control/play the game using gestures as well as button presses. It uses a combination of motion sensing and infrared detection to sense its poses (rotations and translations) in 3D space. The Wii Remote has a 3-axis accelerometer, an infrared camera with an object tracking system and eleven buttons used as input features. In order to provide feedback to the user, the Wii Remote contains four LEDs, a rumble to make the controller vibrate and a speaker. The Wii Remote communicates with the Wii console or with a computer via Bluetooth wireless link, reporting back data at 100 Hz. The reported data contains information about the controller state (acceleration, buttons, infrared camera, etc.). Several studies have been done using the Wii Remote as an interaction device, particularly in the construction of interactive whiteboards, finger tracking systems and control of robots [25].

In order to extract relevant information from the Wii Remote, the motion sensor and the infrared capabilities of the controller were explored. After some experiments, it was concluded that the infrared capabilities of the Wii Remote were not usable. The Wii Remote's infrared sensor offers the possibility to locate infrared light sources in the controller's field of view, but with a limited capacity. The problems arise with the placement of the infrared source in the cell (especially when an object gets between the Wii Remote and the infrared source), the limited viewing angle of the Wii Remote, calibration of the infrared sensor, the limited distance from the Wii Remote to the infrared source that the user should maintain during the demonstration process and detection problems when other infrared sources are around.

Therefore, it was demonstrated that the information provided by the motion sensor would be used to achieve the goals. This motion sensor is a 3-axis accelerometer (ADXL330, Analog Devices, USA), physically rated to measure accelerations over a range of at least +/- 3g, with a sensitivity of 300 mV/g and sensitivity accuracy of 10%.

2.3 Speech recognition

The ASR systems have been used with relative success in the control of industrial robots, even in the presence of surrounding noise. An ASR system similar to that presented in [5] is used, allowing that during the robotic demonstration, the user can use voice commands to act remotely on the robot or on the code generation system. For example, if the user wants to stop the robot motors the command "ROBOT MOTORS OFF" is spoken. Otherwise, if the user wants to generate robot code, for example, a command to move the robot linearly to the current pose, the command "COMPUTER MOVE LINE" is used. It is important to note that each voice command must be identified with a confidence higher than 70%, otherwise it is rejected.

2.4 Force control

The robotic manipulators are often in direct contact with their surrounding environment. For purely positioning tasks such as robotic painting, where the forces of interaction with the environment are negligible, no information about force feedback is required. However, in applications such as polishing, grinding or even in the manipulation of objects, knowledge of the contact forces has a great influence on the quality and robustness of the process.

In the robotic platform presented here, the implemented force control strategy will avoid excessive contact forces between the robot tool and workpiece and, at the same time, detect and avoid collisions during the robot operation. A F/T sensor that measures both force and torque along three perpendicular axes is used, allowing the user to have a better perception of the surrounding environment. The application that manages the cell is continuously receiving feedback from the F/T sensor and if any component of force or torque exceeds a set value $F_{ZA}$ (alert force), a command is sent to make the Wii Remote vibrate (tactile feedback). Moreover, if the value of that component is increased by a percentage $P$ or more, the robot immediately stops (see section 3.3). The $F_{ZA}$ value and percentage $P$ is set by the user and depends on the robot tool, robot speed, workpiece materials, etc.

In order to illustrate the technique, an example is presented. The vacuum suction cup grip attached to the robot wrist makes a vertical approximation to the workpiece (Fig. 4). Analyzing figures 4 and 5, we have that:

(a) Approximation phase, the force $F_Z$ (force component along the Z axis) reflects the weight of the tool.
(b) Beginning of the contact between the tool and the workpiece. $F_Z$ increases rapidly, reaching $F_{ZA}$ (the Wii Remote vibrates).
(c) In this phase, if the user corrects the robot movement the contact force is reduced and the Wii Remote stops vibrating, but, if the robot movement is not corrected, the force limit value $F_{ZL}$ (1) is reached and the robot immediately stops.
(d) The robot tool is in contact with the workpiece with an acceptable contact force.

$$F_{ZL} = F_{ZA}(1 + P), P \in [0, 1] \tag{1}$$

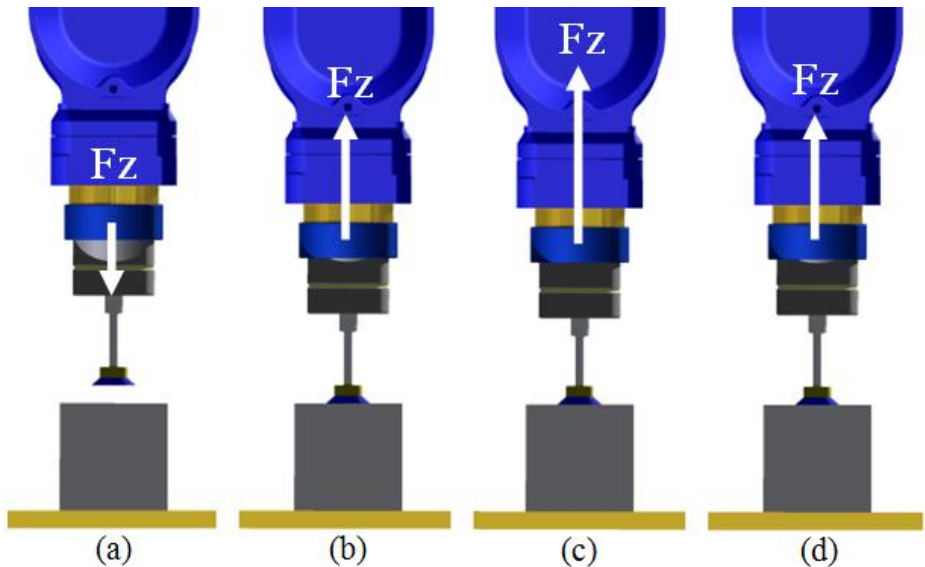

Fig. 4. The various stages of the robot approximation to a workpiece, where the contact forces are highlighted.

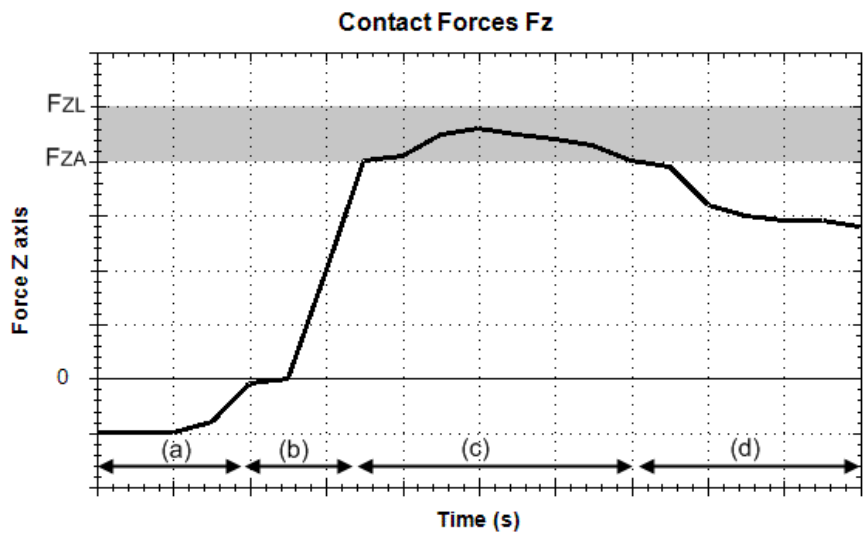

Fig. 5. The evolution of the contact forces $F_z$ between the tool and the workpiece.

2.5 Robot code generation

In the construction of an algorithm to generate code, the keyword is "generalise" and never "particularise". In other words, the algorithm must be prepared to cover a wide range of variations in the process. In this work, the code generation algorithm receives instructions from the identified spoken commands. Thus, during the demonstration the user uses speech to build the robot code step-by-step (write any type of variables, robot commands, etc.), without specific know-how of the native robot programming language. Finally, after finalizing the robotic demonstration task, the user can generate the entire robot program, upload it to the robot controller and run it. If desired, the generated robot code can then be edited and modified.

## 3. Control strategy

### 3.1 Robot control

The robot is controlled remotely via the Ethernet using the MOTOMAN IMOV function that moves the robot linearly according to a specified pose increment $i = [i_1 \ i_2 \ i_3 \ i_4 \ i_5 \ i_6]^T$. The first three $i$ components represent the robot translation along the X, Y, and Z axes, respectively, while the last three $i$ components represent the robot rotation about the X, Y, and Z axes, respectively. These $i$ components have the necessary information to control the robot. It is therefore necessary to identify them by examining the behaviour of the user hand that holds the Wii Remote.

In this system it is completely unnecessary to extract precise displacements or rotations, being only required to know which of the pose increment components must be activated. In a first approach, the robot control strategy was to identify translation movements and rotations of the user hand and, depending on these inputs, small pose increments were continuously sent to the robot. However, it was quickly concluded that this approach was not viable because the robot was constantly halting, presenting a high-level of "vibration". The achieved solution was to send to the robot only one pose increment that will move the robot to the limit of the field of operation.

The robot movement is activated by pressing the Wii Remote B button and making a hand gesture or posture according to the desired robot movement. After this, if a gesture is recognized, the robot starts the movement and when the user releases the B button the robot stops. If the B button is never released, the robot continues the movement up to the limit of its field of operation. If a gesture is not recognized, the robot remains stopped.

### 3.2 Field of operation of the robot – increment calculation

According to the user hand behaviour, the robot is moved to the limit of its field of operation, or more specifically, for a pose close to the limit of the field of operation. The field of operation of a 6-DOF robot manipulator is approximately a volume region bounded by two spherical surfaces. This way, it can be considered that the field of operation of the robot is bounded by two spherical surfaces (2), both with the centre coincident with the zero reference point of the robot, and where $R_{ext}$ and $R_{int}$ are respectively the radius of the external and internal spherical surface.

$$\begin{cases} x^2 + y^2 + z^2 \leq R_{ext}^2 \\ x^2 + y^2 + z^2 \geq R_{int}^2 \end{cases} \quad (2)$$

Before starting any robot movement, the "current" robot position $(x_r, y_r, z_r)$ is acquired. In order to calculate the pose increment $i$, firstly it is necessary to calculate the increment components which must be activated. This is done by referring to the Wii Remote acceleration values that will define the robot movement direction $a = (a_x, a_y, a_z - 1)$ (see section 4). From the vector $a$, a unitary vector $\hat{u}$ can be defined with the same direction, the direction of the robot movement. This vector $\hat{u}$, in conjugation with the "current" robot position point $(x_r, y_r, z_r)$, will be used to achieve a straight line (3) that will intersect the external spherical surface at two points (Fig. 6). In a first approach, it is considered that only the external spherical surface limits the robot field of operation.

$$(x, y, z) = (x_r, y_r, z_r) + k\hat{u}, k \in \mathbb{R} \tag{3}$$

From (2) and (3):

$$(x_r + k\hat{u}_1)^2 + (y_r + k\hat{u}_2)^2 + (z_r + k\hat{u}_3)^2 = R_{ext}^2 \tag{4}$$

Extracting $k$ from (4), and considering only the positive value of $k$ (vector $\hat{u}$ direction), the distance from the "current" robot position to the external spherical surface point (robot increment) is (5).

$$(x, y, z) = (i_1, i_2, i_3) = k(\hat{u}_1, \hat{u}_2, \hat{u}_3), k \in \mathbb{R}^+ \tag{5}$$

The value of $k$ depends on the volume reached by the robot. In the work presented here, using a MOTOMAN HP6 robot, the value of $k$ is limited to the interval [0, 2012].

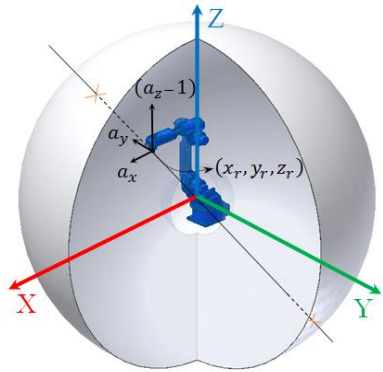

Fig. 6. The two spherical surfaces that define the robot field of operation. The "current" robot point and the acceleration vector components that will define the robot movement direction are represented in the figure.

Thus, in terms of robot translation movements, the pose increment is $i = [i_1 \quad i_2 \quad i_3 \quad 0 \quad 0 \quad 0]^T$. Note that, for example, if it is found (in the gesture recognition phase) that the robot should be moved along the X axis in the negative direction, the vector $\hat{u}$ becomes $(-1,0,0)$, and then $i = [i_1 \quad 0 \quad 0 \quad 0 \quad 0 \quad 0]^T$.

An analogue approach was employed to obtain $i$ when the robot field of operation is limited by the internal spherical surface. In this case, if $k$ has no value (impossible to calculate), it means that the straight line does not intercept the internal spherical surface and it is the external spherical surface that limits the robot field of operation.

In terms of rotations, since we know the robot rotation limit values and the "current" robot pose, it is easy to obtain the increments $i_4$, $i_5$, and $i_6$.

3.3 Security systems

When a human interacts directly with a robot in a co-worker scenario, the security systems present in the robotic cell should have a high-level of robustness, in order to avoid accidents. A system was implemented in our robotic cell a system that is continually receiving data from the Wii Remote (via Bluetooth) and if the communication fails, the robot immediately stops. An independent system was implemented an independent system that actuates directly in a low-level of the control hierarchy, stopping the robot if any problem occurs. This system operates independently from the software that is running on the robot or computer.

Since the Wii Remote communicates via Bluetooth, it is important to discuss the reliability of this technology. Investigations and practical tests have proven Bluetooth's reliability, but caution should be exercised when installing Bluetooth products. Important issues are interference with other Bluetooth nodes, radio standards, and sources of radiation (industrial equipment or commercial devices like microwave ovens). Thus, considering the high demands for safety and real-time performance in industry, this technology must be used with care.

## 4. Posture and gesture recognition

4.1 Modes of operation

The developed system has two distinct modes of operation that the user can select during the demonstration phase. In the first mode, the robot moves along the X, Y, and Z axes separately, while in the other mode the robot can move along the three axes at the same time. In terms of rotations, in both cases, the rotation around each of the three axes is done separately, an axis at a time.

The accelerations extracted from the 3-axis accelerometer $(a_x, a_y, a_z)$ will be used to detect the user hand gestures and postures. When the Wii Remote is operating in a dynamic way, the gravity components will appear mixed with the inertial components of acceleration. In order to prevent this situation, when the user makes gestures, the Wii Remote must be kept horizontal (Fig. 7). Thus, it is known that the force of gravity acts along the Z axis. For example, to move the robot in the X direction, the user should move the Wii Remote along the X axis, keeping it in the horizontal.

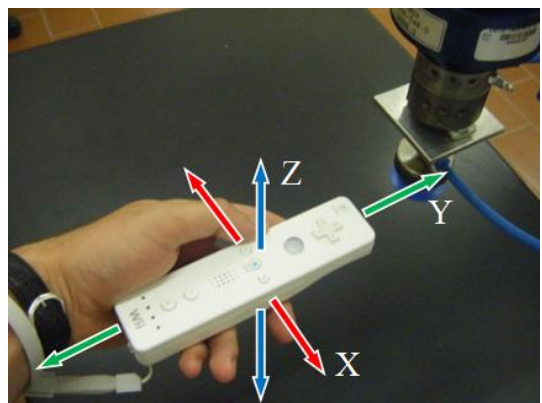

Fig. 7. The developed system can recognize six different gestures (X+, X-, Y+, Y-, Z+, and Z-), it is necessary to take full control of the robot in terms of robot translations. In both movements the Wii Remote is held horizontally.

## 4.2 Recognition of gestures and postures

In accelerometer-based gesture recognition, the signal patterns produced by the accelerometers are used in generating models that allow the recognition of different gestures. Moving the Wii Remote over each one of the three axes (in both directions), we can extract $(a_x, a_y, a_z)$ for each of the six different gestures necessary to take full control of the robot translations (X+, X-, Y+, Y-, Z+, and Z-). When the Wii Remote is moved in the positive X direction (X+) (Fig. 8), initially the value of acceleration $a_x$ increases because the hand begins to move and then, when the hand begins to slow the positive value of $a_x$ is converted to a negative value. This point ($a_x = 0$) marks the point of maximum speed. The acceleration $a_y$ remains near to zero, and $a_z$ remains near to one because the Wii Remote is held horizontally. A similar reasoning can be done to the other gestures (X-, Y+, Y-, Z+, and Z-).

To interpret the acceleration values and recognize the hand movements, a statistical approach was used. For each of the six gestures is calculated the arithmetic mean of the accelerations measured in the training phase, $(\overline{a_x}, \overline{a_y}, \overline{a_z})$ (from the beginning of the movement to the first point of zero acceleration (maximum speed)). After this, the standard deviation is used to measure how widely spread the acceleration values are from each mean $(\sigma_x, \sigma_y, \sigma_z)$. This way, in the training phase a range of acceleration values is established, which define each of gestures. During the robotic demonstration phase, a gesture is recognized when:

$$a_x \in [\overline{a_x} - \sigma_x, \overline{a_x} + \sigma_x]$$
$$a_y \in [\overline{a_y} - \sigma_y, \overline{a_y} + \sigma_y]$$
$$a_z \in [\overline{a_z} - \sigma_z, \overline{a_z} + \sigma_z]$$

Where $(a_x, a_y, a_z)$ are means of the acceleration values measured during the robotic demonstration phase (again from the beginning of the movement to the first point of zero acceleration (maximum speed)). However, under this approach, the robot begins to move after the system recognizes the gesture of the hand, showing a considerable delay from the press of the B button to when the robot starts to move. This delay is due to several acceleration measurements made during the execution of the gesture.

Thus, an alternative methodology was investigated. To achieve the goals, this time delay is not acceptable. The aim is that the robot starts the movement almost at the same time as the user starts the hand movement and presses the B button. To do this, immediately after the user starts the hand movement and presses the B button, the system extracts the acceleration values from the Wii Remote, identifies the gesture and sends a command to move the robot. It is quite complicated to do this in few milliseconds because it is necessary to make several measurements of acceleration over time. This problem can be solved by making fewer measurements of accelerations. However, if the number of measured acceleration is reduced, it is more difficult recognize a gesture and the recognition rate becomes low. The goal is to achieve a compromise between the time delay, and the achieved recognition rate. After some experiments were done to evaluate the system response time, it was decided to extract from the accelerometer only the first four measurements of acceleration that constitute an input pattern. A gesture is identified by comparing the extracted acceleration values with the values acquired in the training phase, following the same statistical approach outlined above. This alternative methodology was then developed using ANNs.

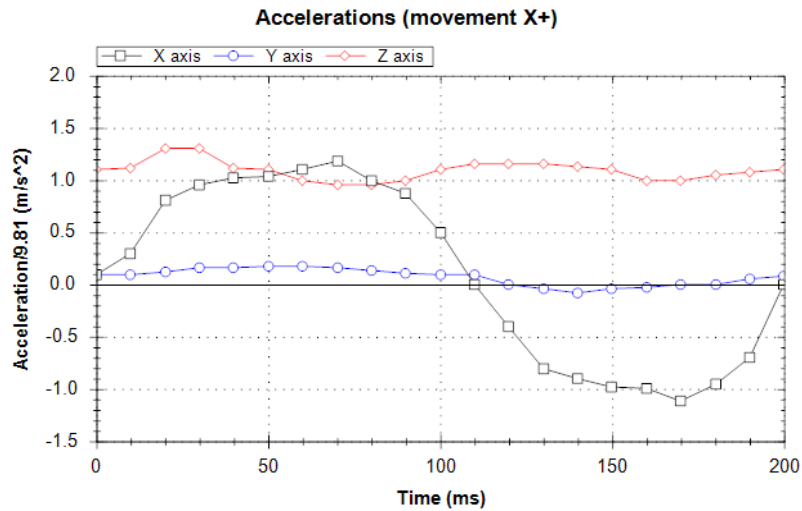

Fig. 8. The measured accelerations, when the Wii Remote is moved along the X axis in the positive direction (X+).

In the second mode of operation, the robot is moved linearly along the direction that the user hand demonstrates, in other words, the vector of accelerations $a = (a_x, a_y, a_z - 1)$ will directly define the robot movement direction. The vector $a$ is extracted immediately after the user starts the hand movement and presses the B button (this vector is defined as a mean of the four first measurements of acceleration). The third component of $a$ is $(a_z - 1)$ since the Wii Remote is held horizontally, reporting an acceleration along the Z axis.

Besides the robot translations, the robot control architecture needs also to have as input six different robot rotations (Rx+, Rx-, Ry+, Ry-, Rz+, and Rz-). If the Wii Remote is in free fall, it will report zero acceleration. But if the Wii Remote is held horizontally with the "A" button facing up (Fig. 9-A), it will report an acceleration along the Z axis, the acceleration $g$ due to gravity, approximately $9.8 m/s^2$. Thus, even when the user is not accelerating the Wii Remote, a static measurement can determine the rotation of the Wii Remote (hand posture recognition).

Analyzing figure 9-A, when the Wii Remote is held horizontally, it will report an acceleration $g$ along the Z axis in the positive direction; $a_z \approx g$, $a_x \approx 0$, and $a_y \approx 0$. But when the Wii Remote is rotated around the Y axis (Fig. 9-B); $a_x \approx g$, $a_y \approx 0$, and $a_z \approx 0$. On the contrary, when the Wii Remote is rotated around the Y axis in the reverse direction (Fig. 9-C); $a_x \approx -g$, $a_y \approx 0$, and $a_z \approx 0$. A similar approach detects rotations around the X axis (Fig. 9-D and 9-E). In the detection of gestures (robot translations without time delay), a range of acceleration values is established which define each of the postures (Rx+, Rx-, Ry+, and Ry-). However, in terms of rotation around the Z axis (Fig. 9-F, 9-G), nothing can be concluded as in both cases the gravity is along the Z axis. To solve this problem, an ANN was used to detect rotation movements around the Z axis, and this method was then also used for the other two axes.

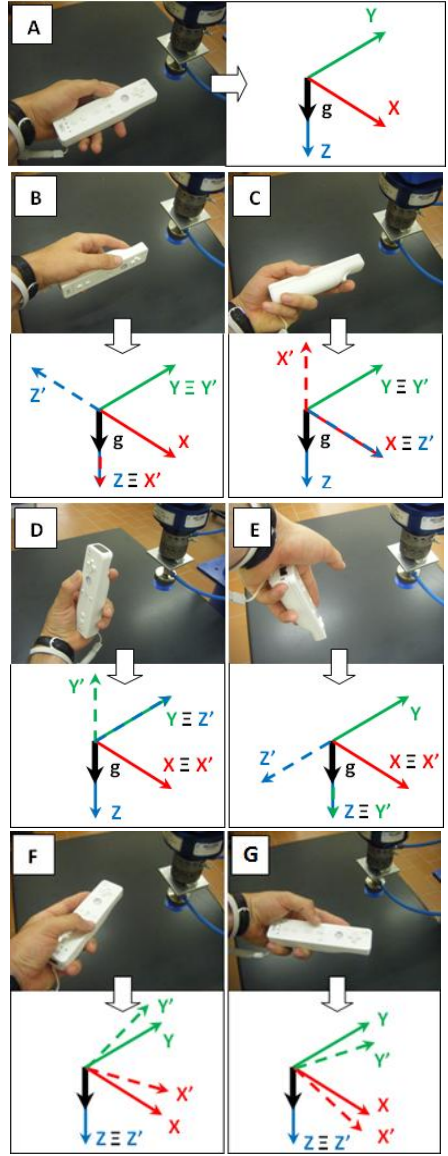

Fig. 9. A-No rotation, B-Rotation around the Y axis in the negative direction (Ry-), C – (Ry+), D – (Rx-), E – (Rx+), F – (Rz+), G – (Rz-).

### 4.2.1 Neural Networks

The ANNs have been applied in the recognition of gestures with success, especially because they are simple to use and efficient in the recognition of patterns. Moreover, the ANNs present good learning capabilities, such as, the ability to learn from experience and to generalize from examples to correctly respond to new data.

In order to detect and identify gestures and postures, an ANN trained with a backpropagation algorithm was implemented. The input signals (acceleration data) are represented by a vector $x = (x_1, \ldots, x_n)$, and the output from a neuron $i$ is given by (6), where $x_j$ is the output of neuron $j$, $w_{ij}$ is the weight of the link from neuron $j$ to neuron $i$, $\theta_i$ is the bias of neuron $i$, and $F$ is the activation function.

$$x_i = F\left(\sum_j w_{ij} \cdot x_j + \theta_i\right) \tag{6}$$

It is now necessary to find the weights of the network. The backpropagation algorithm is used as a learning algorithm to determine the weights of a network, in other words, the method adjust the weights to minimize the errors. These errors can be determined from the input neurons (training set of several patterns) and the desired output vector. The error is achieved comparing the desired output (obtained in the training phase) with the actual output. This methodology presupposes that gestures trained for performing certain functions are always repeated as they were trained.

### 4.2.2 Methodology

A specific gesture or posture is recognized by a three-layer ANN (Fig. 10). The number of neurons was twelve for the input layer, twenty for the hidden layer and twelve for the output layer. Twelve neurons in the input layer encode each gesture, four measurements of acceleration and each with three components of acceleration. Twenty neurons were used in the hidden layer because after several experiments it was concluded that this solution presents a compromise between the computational time required to train the system and the recognition rate. Finally, the twelve neurons in the output layer correspond to each different gesture/posture.

In the training phase, the user should train the system, demonstrating the hand movement for each gesture and posture several times (X+, X-, Y+, Y-, Z+, Z-, Rx+, Rx-, Ry+, Ry-, Rz+ and Rz-). The acceleration values representing each gesture/posture are given to the input neurons (learning patterns), and at the same time, one of the output neurons corresponding to the presented gesture or posture is set to 1.

Each neuron of the output layer outputs the recognition result, a numerical value between 0 and 1 (sigmoid function). If the output value is larger than or equal to 0.5, it means that the neuron detected the gesture or posture. A gesture/posture is recognized with a confidence higher than 80%. If a gesture is classified correctly and if the output neuron related to this gesture has a value near to 1, the confidence is high. If a gesture is classified correctly, but the output value of a specific neuron is not so close to the training data, the confidence is low.

Finally, to define the robot increment, the recognized gestures are then transformed in the vector $\hat{u}$, for example, if is detected the movement (Y+), $\hat{u} = (0,1,0)$.

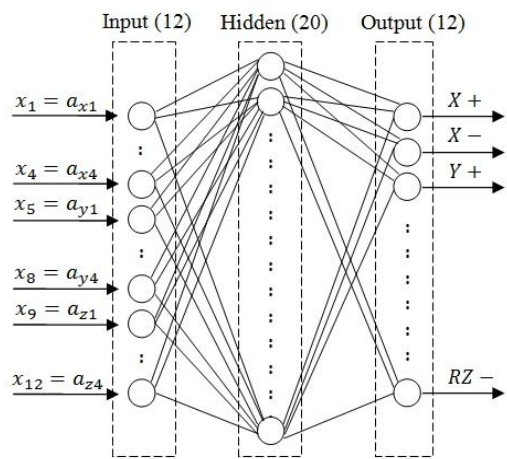

Fig. 10. Structure of the ANN used for recognition of gestures and postures.

4.2.3 Recognition rate and discussion of results

In this section, the results obtained through several experiments are presented, discussed, and compared with other approaches.

After training the system, several tests were conducted to achieve the recognition rate for each gesture and implemented method. The first method presented here, statistical approach with a time delay is named Method 1, the second, statistical approach without time delay is named Method 2, and the third, ANN approach without time delay is named Method 3. The tests were conducted with four participants, and each one performed each gesture/posture 100 times. The participants were two experienced users that trained the system before performing the tests (P1 and P2) and two first-time users that used the system trained by the other users (P3 and P4). Table I presents the recognition rate for each gesture and posture, varying the method and the participants. The recognition rate is presented as the mean for the participants P1 and P2, and the same for the participants P3 and P4. These results are obtained with 30 patterns taught to the network in the training phase. For Method 3, the ANN parameters are presented in table II.

Table I Recognition rate for different gestures and postures.

| Gesture or posture | Recognition rate (%) | | | | | |
| --- | --- | --- | --- | --- | --- | --- |
| | Method 1 | | Method 2 | | Method 3 | |
| | P1 & P2 | P3 & P4 | P1 & P2 | P3 & P4 | P1 & P2 | P3 & P4 |
| X+ | 99 | 80 | 94 | 79 | 97 | 83 |
| X- | 98 | 77 | 93 | 75 | 95 | 77 |
| Y+ | 99 | 78 | 92 | 74 | 96 | 78 |
| Y- | 97 | 78 | 93 | 72 | 93 | 76 |
| Z+ | 98 | 90 | 95 | 77 | 98 | 81 |
| Z- | 98 | 84 | 90 | 69 | 96 | 75 |
| RX+ | 100 | 93 | 100 | 91 | 100 | 96 |
| RX- | 100 | 96 | 100 | 94 | 100 | 95 |
| RY+ | 100 | 97 | 100 | 90 | 100 | 92 |
| RY- | 100 | 98 | 99 | 89 | 100 | 95 |
| RZ+ | 95 | 71 | 89 | 68 | 92 | 71 |
| RZ- | 93 | 67 | 86 | 65 | 90 | 69 |
| **Mean** | **98** | **84** | **94** | **79** | **96** | **82** |

Table II ANNs parameters and results.

| | |
| --- | --- |
| **Activation function** | Sigmoid function |
| **Training cycles** | 100000 |
| **Number of hidden neurons** | 20 |
| **Learning rate** | 0.25 |
| **Momentum** | 0.1 |
| **Computer processor** | Intel core 2 duo T5600 |
| **Computer RAM** | 1 GB |
| **Computational time** | 34 Minutes |
| **Training time** | 6 Minutes |

In Method 1, the experiments showed a recognition rate of 98% for the participants P1 and P2, and 84% for the participants P3 and P4. The Method 2 and 3 do not present such a good average of correctly recognized gestures, but in compensation do not have the time delay, which is crucial in the system. Method 3 (ANN approach without time delay) was shown to be the best solution, with a recognition rate of 96% for the participants P1 and P2, and 82% for the participants P3 and P4. For P1 and P2, even the lowest recognition rate of the gesture RZ- is 90%. It was concluded that the participants P1 and P2 present better results than the participants P3 and P4, demonstrating the necessity for each user to train the system before using it. However, the aim is that the time spent in the training should be minimal. The users do not want to spend time demonstrating gestures and postures to the system, but at the same time it is necessary to keep a compromise with the recognition rate.

The recognition rate of the output depends a lot on the samples provided during the training phase (learning patterns). Given that Method 3 proved to be a good solution, several experiments were made to test its performance when the number of samples given to the ANN in the training phase is changed. In table III are presented the results, keeping the parameters of the ANN described in table II. It was found that the model requires 30 samples per gesture to achieve an accuracy of about 96%, but if the number of learning patterns is increased to 60 or 70, the recognition rate is improved but not significantly. Giving 30 patterns to the system the user takes 6 minutes in the training phase, and with 60 patterns the user takes 11 minutes.

Each different user can train the system using an intuitive interface (Fig. 11). Moreover, the system can be trained to recognize other gestures than those already mentioned, for example, a gesture to activate a specific robot output signal. The participants commented that it is easy and intuitive use gestures for controlling an industrial robot.

The recognition rate obtained by our system presents similar results to other systems that use ANN to recognize gestures. An ANN-based approach using a data glove as input device presents a recognition rate of 98% [14]. A vision-based gesture recognition system for hand gesture recognition achieved a recognition rate of over 90% [15]. Another study using an accelerometer-based input device to operate televisions presented a recognition rate higher than 97% [9].

**Table III** Effect of the number of learning patterns in the recognition rate.

| Number of learning patterns | Recognition rate (%) | |
|---|---|---|
| | P1 & P2 | P3 & P4 |
| 20 | 94 | 79 |
| 30 | 96 | 82 |
| 60 | 97 | 84 |
| 70 | 97 | 85 |

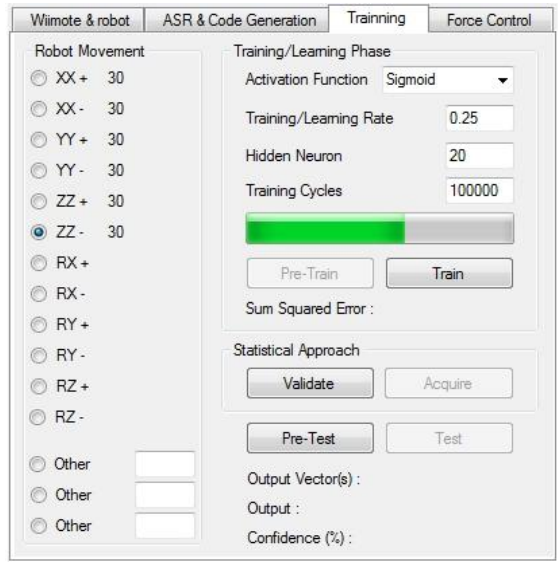

Fig. 11. Application interface for the training phase.

## 5. Practical tests and results

To assess the performance of our system, different practical tests were performed (using Method 3). These tests include a robotic pick-and-place operation, write on a sheet of paper, and move the robot to different poses in the presence of obstacles (Fig. 12). Both practical tests were made in a non-controlled environment and the results obtained were very promising, showing that an untrained user can generate a robot program for a specific task, quickly and in a natural way [27]. Other advantages of the system are:

1) Short set-up time after the training phase.
2) Given the current state of the art in this area, the system presents a good average of recognized gestures and speech.
3) In terms of accuracy of the robot-controlled movements, since the user can control the robot speed, each robot movement can be controlled with a high-level of accuracy. In practice, when the robot is near to the workpiece or obstacle, the robot speed is reduced.
4) The force control helps in the positioning of the robot, avoids excessive contact forces between the robot tool and workpiece, and at the same time detects and avoids collisions during the robot operation.
5) Usability of the system for non-expert robot programmers.
6) Offers the possibility to control and program different robots using the same basic gestures.

The drawbacks are:

1) Complexity of the system (communications, security).
2) Due to the fact that the Wii Remote and the headset communicate with the computer via Bluetooth, the user should remain close to the computer (must be used within 7 meters).

3) Time spent in the training phase. Each user must train the system to recognize his or her particular hand behaviours.
4) Reliability of Bluetooth devices.

However, in terms of robot control, when compared with other common input devices, especially the teach pendant, this approach using the Wii Remote is more intuitive and easy to use, besides offering the possibility to control a robot by wireless means.

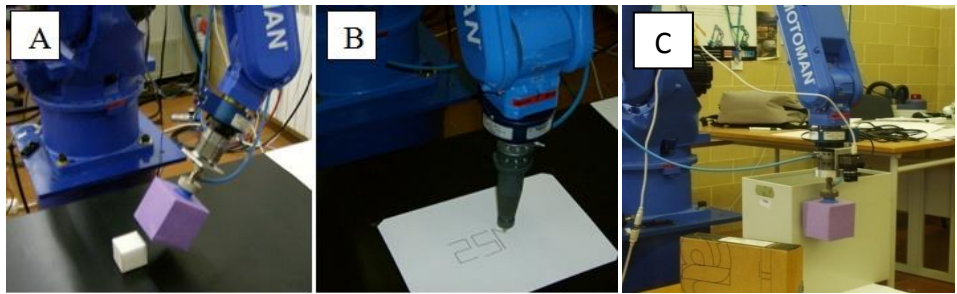

Fig. 12. A – Robotic pick-and-place operation, B – Robot writing letters, C – Robot moving in the presence of obstacles.

The system response time for Method 3 was evaluated. It is difficult to measure the response time, assuming that it is fractions of a second. Due to the complexity of the system, a camera recorder was used to film the robot and the user holding the Wii Remote. Analyzing the captured film and counting the frames from the press of the Wii Remote B button to when the robot started to move, the system response time was measured. After several experiments, it was concluded that the response time of our system is 140 milliseconds, which is quite reasonable.

5.1 Comparison with a manual guidance system

The performance of the system developed (Method 3) was compared with a similar system that instead of gestures uses a manual guidance system based on a force control strategy to move/guide the robot in space [3], [28] and [29], (Fig. 13). Both systems are intuitive and easy to use, however, to do the same robotic task the manual guidance system takes less time. A test was conducted, where the robot wrote three letters (S, M and E). Our system takes 2 minutes and 15 seconds to write the three letters with a robot speed of 75 mm/s, while the manual guidance system takes 30% less time. In addition to the above, the manual guidance system presents better robustness than our system that sometimes does not recognizes the hand postures and gestures. Moving a robot through manual guidance gives the user a feeling of greater control and involvement with the robot that the user can not have when using the Wii Remote. However, the manual guidance system also presents some drawbacks that limit its use, such as the high price, and the fact that this system can only be incorporated in ABB robots. The aim is that with further development, our method can be more practical than or equally practical to manual guidance.

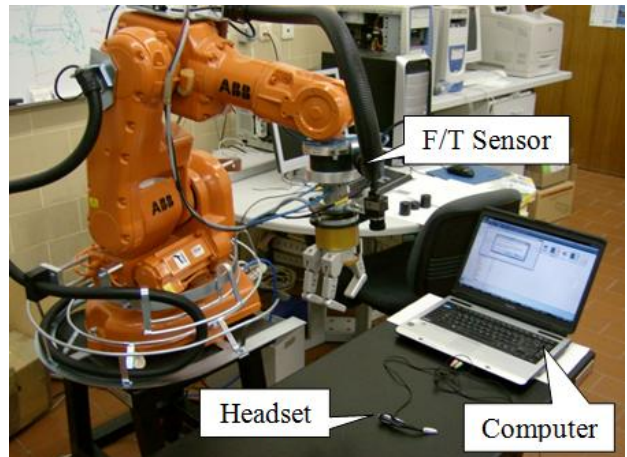

Fig. 13. Due to their force control system, the ABB robot (IRB 140 equipped with the IRC5 controller, ABB, Switzerland) can be guided manually by the user.

## 6. Future research

The recognition rate for gestures and postures should be high (nearly 100%). This high recognition rate is required because a low recognition rate may cause the users to abandon the method. Future work will seek to improve the recognition rate of gestures and postures. Another interesting aspect would be to achieve precise displacements from the measured accelerations, despite the risks of systematic errors accumulated in the double integration phase. Finally, a long term goal is continuous gesture recognition.

Since the input devices containing accelerometers are cheap, developments in accelerometer-based gesture recognition is increasing significantly. This ability to detect gestures is useful not only in the robotics field but also for interacting with other different kinds of devices.

## 7. Conclusion and discussion

Due to the growing demand for natural HMIs and robot intuitive programming platforms, a robotic system that allows users to teach and program an industrial robot using gestures and speech was proposed. It was shown how a novel game input device with motion and infrared sensing capabilities could be integrated in a robotic cell to obtain a high-level programming environment.

Special emphasis was placed on the recognition of gestures and postures in a non-controlled environment, where the motion sensing capabilities of the Wii Remote were used for accelerometer-based gesture recognition. Experiments showed that an approach using ANNs is a good solution to achieve a reasonable gesture and posture recognition rate up to 96%. Moreover, the system presents an acceptable response time, 140 milliseconds. This is the time delay from the moment the user starts to perform the gesture until the robot starts to move.

The Wii Remote is an off-the-shelf product, available in the market at a low price. Using this and other standard technologies means that future innovation will be faster and less expensive. Finally, most of users found it natural and intuitive to use gestures for controlling an industrial robot.

## 8. Acknowledgment

This work was supported in part by the European Commission's Sixth Framework Program under grant no. 011838 as part of the Integrated Project SMErobot™, and the Portuguese Foundation for Science and Technology (FCT) (SFRH/BD/39218/2007). The authors want also to acknowledge the help of the Portuguese Office of the Microsoft Language Development Centre, especially Professor Miguel Salles Dias, for their support with the Microsoft ASR and TTS engines and related APIs.